\documentclass{article}
\usepackage{spconf,amsmath,graphicx}
\usepackage{booktabs}
\usepackage{multirow}
\usepackage{enumitem}
\usepackage{bbding}
\usepackage{amssymb}
\usepackage[cmyk]{xcolor}
\setlist[itemize]{leftmargin=*}
\usepackage[ruled,linesnumbered]{algorithm2e}
\usepackage{algorithmic}
\usepackage{subfigure} 

\title{Completely Heterogeneous Federated Learning} 
\name{Chang Liu, Yuwen Yang, Xun Cai, Yue Ding, Hongtao Lu\thanks{$*$Corresponding author: Hongtao Lu, htlu@sjtu.edu.cn }$^{*}$}

\address{Department of computer science and engineering, Shanghai Jiao Tong University \\ 
    \{isonomialiu, youngfish, caixun, dingyue, htlu\}@sjtu.edu.cn
}

\begin{document}

\maketitle
\begin{abstract}
Federated learning (FL) faces three major difficulties:  cross-domain, heterogeneous models, and non-\textit{i.i.d.} labels scenarios. Existing FL methods fail to handle the above three constraints at the same time, and the level of privacy protection needs to be lowered (\textit{e.g.}, the model architecture and data category distribution can be shared). In this work, we propose the challenging ``completely heterogeneous'' scenario in FL, which refers to that each client will not expose any private information including feature space, model architecture, and label distribution. We then devise an FL framework based on parameter decoupling and data-free knowledge distillation to solve the problem. Experiments show that our proposed method achieves high performance in completely heterogeneous scenarios where other approaches fail.

\end{abstract}

\begin{figure*}[h]
   \centering 
    \subfigure[Completely Heterogeneous FL]{
        \label{Fig.HFL}
        \includegraphics[width=0.375\linewidth]{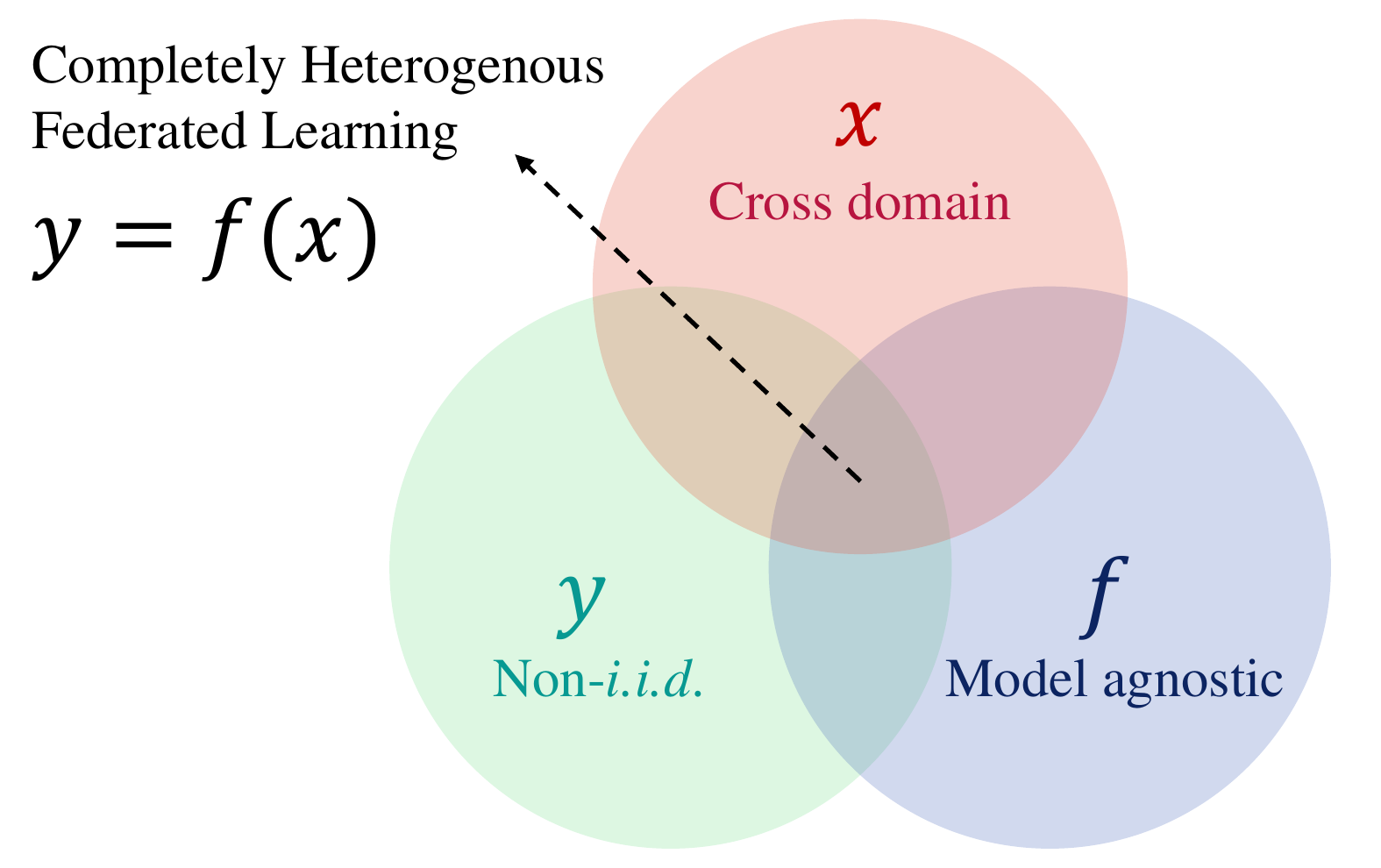}
        }
    \subfigure[Our Method]{
        \label{Fig.method}
        \includegraphics[width=0.6 \linewidth]{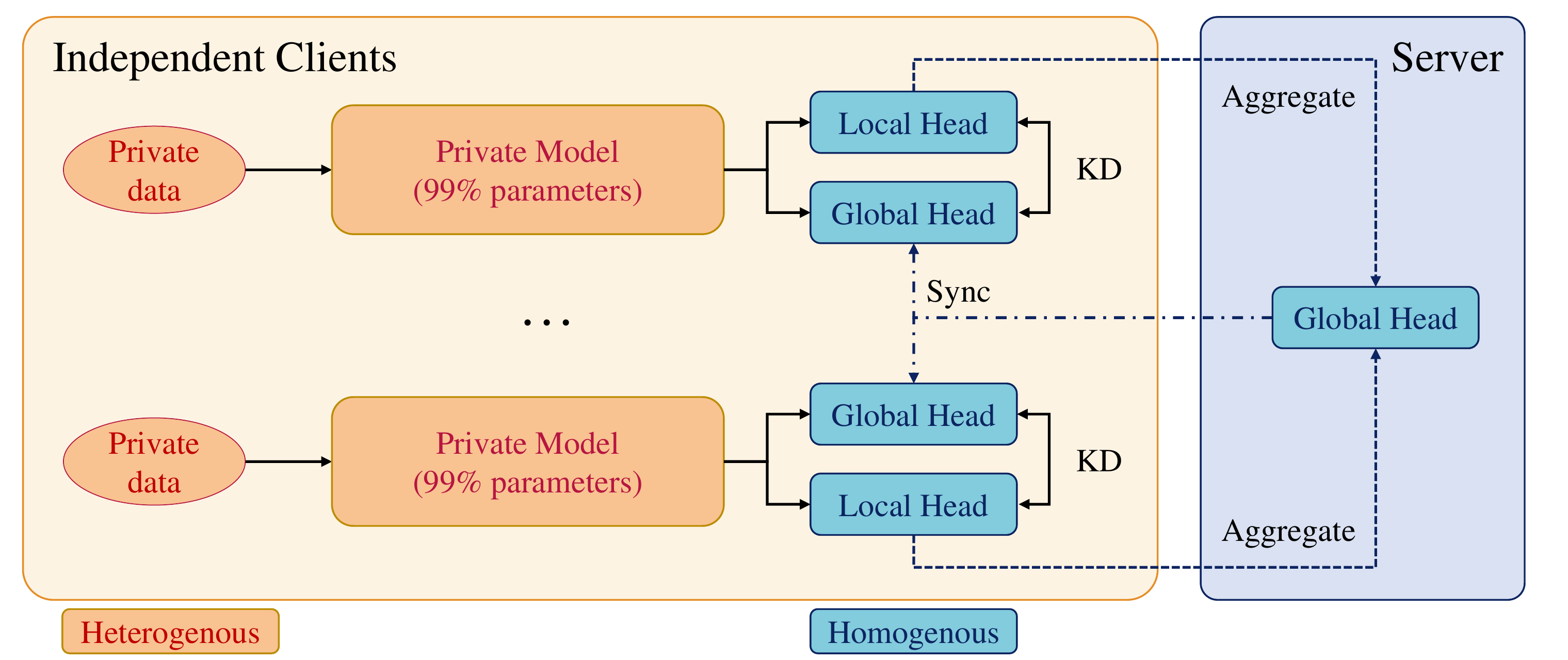}
        }
    \caption{Completely heterogeneous FL framework. (a) Definition of completely heterogeneous FL (b) We propose the parameter decoupling and data-free KD to make completely heterogeneous FL possible.}
    \label{Fig.res}
    \centering 
\end{figure*}

\section{Introduction}
\label{sec:intro}

Federated learning (FL) stems from the need for privacy protection, enabling multiple clients to collaboratively learn a shared model $y=f(x)$ without collecting data from local devices \cite{1stFed,mcmahanCommunicationEfficientLearningDeep2017}. 
The current FL scenarios are limited to requiring participants to meet the following constraints: The \textbf{data} ($x$) should belong to the same feature space (domain). The \textbf{outputs} (y) is preferably independently and identically distributed (\textit{i.i.d.}) across different participants. The same \textbf{model} ($f$) architecture is needed because FL is mostly based on model-level parameter aggregation \cite{yangFederatedMachineLearning2019}. Meanwhile, it is difficult to satisfy all the three above constraints at the same time in FL in real-world scenarios.
Data from various clients are often extracted with different features or from different domains, especially tabular data. The data distribution of different clients is often non-\textit{i.i.d.}, and these specific distributions are usually private and sensitive, such as the types of target customers of major enterprises. In addition, each client has different computing resources, and the model is differentiated and private. This puts FL participants in a dilemma of either abandoning FL or having to compromise on privacy and performance, which hinders the expansion of FL application. 

One of the above three challenges can be addressed individually by existing FL methods. Personalized FL (PFL) is a good paradigm to address the non-\textit{i.i.d.} scenario \cite{kairouzAdvancesOpenProblems2021a}. But PFL requires obtaining statistics on private data or model \cite{liFedH2LFederatedLearning2021}, or training data generator on the server, which implicitly assumes the same data domain or the consistent model design \cite{tanPersonalizedFederatedLearning2022}. Model agnostic FL methods \cite{liFedMDHeterogenousFederated2019, Huang2022FewShotMA, zhangFedZKTZeroShotKnowledge2021} propose to train on a public dataset to enable the federation of heterogeneous models. Public datasets still mean the same feature domain and distribution. FedHeNN \cite{makhijaArchitectureAgnosticFederated2022} train heterogeneous models via transmitting internal embedding to server, which brings high complexity and leakage risk.

In this paper, we aim to address this difficult FL scenario where the feature domain of $x$, the distribution of $y$, and model $f$ are \textbf{completely heterogeneous}. In other words, we expect all the participants no longer need to be constrained by these factors, allowing more flexibility for FL. 
 
\textbf{Contributions.} The contributions of our paper include:
\begin{itemize}
\setlength{\itemsep}{0pt}
\setlength{\parsep}{0pt}
\setlength{\parskip}{0pt}
    \item We extend FL to the ``completely heterogeneous'' scenario, which means that the feature space, label distribution of data, and clients' models are all unknown.
    \item We propose a parameter decoupling strategy, it shows that only a very small portion of parameters (less than 5\%) need to be fixed to ensure the privacy of most model designs. A data-free knowledge distillation is proposed to enable KD-based FL without any shared data.
    \item We conduct fair experiments on image and tabular datasets to demonstrate that our method can better cope with completely heterogeneous scenarios than existing methods.
\end{itemize}

\section{Related works}
 Federated learning is frequently divided into three categories \cite{yangFederatedMachineLearning2019}: horizontal federated learning, vertical federated learning, and federated transfer learning based on the overlap of feature space and sample ID space. 
 The most classical federated average algorithm \cite{mcmahanCommunicationEfficientLearningDeep2017} can be used to update the model in horizontal FL because the feature space of each participant's dataset is largely the same. In vertical FL, however, only a few participants have the complete label information. The optimization process requires frequent private set intersection to align the samples. Federated transfer learning has less feature and sample space overlap between participants and needs domain adaptation for cross-participant model learning. In contrast, our method does not require assumptions on the overlap between the feature and the sample ID space across participants. Our method also requires no extra alignment operations on the model parameters during training. In other words, we can handle both homogeneous and heterogeneous conditions for datasets and model architectures. 

To achieve the ``completely heterogeneous'' FL, we introduce parameter decoupling \cite{arivazhaganFederatedLearningPersonalization2019,collinsExploitingSharedRepresentations2021,PillutlaMMRS022,PartialFed,FedPS} by reserving non-shared model parameters for potentially heterogeneous datasets. At the same time, to hide most of the model architecture and parameters, we combine parameter decoupling with the knowledge distillation \cite{hintonDistillingKnowledgeNeural2015} algorithm often employed for heterogeneous model learning \cite{liFedMDHeterogenousFederated2019,zhangFedZKTZeroShotKnowledge2021,  choHeterogeneousEnsembleKnowledge2022a, linEnsembleDistillationRobust2020}. Existing parameter decoupling methods require explicit knowledge of the model architecture and aggregation of most of the model parameters, which does not match the requirements of this scenario. Besides, existing KD-based methods need to share datasets or transfer embeddings between clients, which is unacceptable in some real-world cases.

\section{SYSTEM MODEL}
We firstly define the ``completely heterogeneous'' scenario. Consider an FL classification task with $K$ participating clients. Each client has a model $f$ with heterogeneous paradigms, \textit{i.e.,} $f_k \in \{ f_1(\cdot), f_2(\cdot), \cdots, f_K(\cdot)\}$. We assume that they have heterogeneous datasets $\mathcal{D}^k = \{\boldsymbol{X}_k,\boldsymbol{Y}_k\}$, which means the different feature space $\boldsymbol{X}_k$ and non-\textit{i.i.d.} distributed $\boldsymbol{Y}_k$. For the $k$-th client, 
\begin{equation}
\label{eq:forward}
    \hat{y}_k = f_k(x_k),
\end{equation}

where $x_k \in \mathbb{X}_k$ denotes an input data sample from the $k$-th client's feature space $\mathbb{X}_k$, $f_k$ is the private model, and $\hat{y}$ is the predicted output of the given sample $x_k$. Then completely heterogeneous FL aims to optimize the following formula:
\begin{equation}
\label{eq:opt}
\begin{aligned}
& \min _{\mathbf{w}_k \in \mathbb{F}_k} \frac{1}{K} \sum_{k=1}^K \mathcal{L}_k(\mathbf{w}_k), \\
\text{where \;} & \mathcal{L}_k = \frac{1}{N_k} \sum_{n=1}^{N_k} \mathcal{L}_{\text{CE}}(\mathbf{w}_k; x_{k}^{n}, y_{k}^{n}), 
\end{aligned}
\end{equation}
where $\mathbf{w}_k \in \mathbb{F}_k$ means the parameters of client $k$'s model $f_k$ obeying its specific model design space $\mathbb{F}_k$, $\mathcal{L}_k$ means the loss of the $k$-th client, $\mathcal{L}_{\text{CE}}$ denotes the Cross-Entropy loss.
Note that current FL algorithms based on gradient aggregation or parameter averaging will fail if $i \neq j$ and $\mathbb{F}_i\neq \mathbb{F}_j$. The feature space are also different (and private), \textit{i.e.,} $\mathbb{X}_i\neq \mathbb{X}_j$, so that the existing model agnostic FL methods \cite{liFedMDHeterogenousFederated2019, Huang2022FewShotMA,zhangFedZKTZeroShotKnowledge2021} requires the same shared dataset will also fail.

\section{METHODOLOGY}
\subsection{Parameter decoupling}
In order to solve the problem of model heterogeneity, we need to make the following compromises on assumptions:
\begin{enumerate}
\setlength{\itemsep}{0pt}
\setlength{\parsep}{0pt}
\setlength{\parskip}{0pt}
    \item Clients can mutually agree to align the full set of of label space in all datasets.
    \item Clients can agree on the same length $E$ for all models' output embedding.
\end{enumerate}
 We take a multi-classification task with $C$ classes as an example. Assumption 1 means that although the $i$-th client may not have a sample of a certain category, it will still fill the corresponding position with zero in a $1\times C$ one-hot vector. Assumption 2 means that all models can have homogeneous head modules $h_k(\cdot)$, which means parameters of all $h_k(\cdot)$ are in the same shape. In this way, model $k$'s parameters $\mathbf{w}_k$ can be decoupled into two parts, \textit{i.e.,} the head parameters $\mathbf{w}_{h_k}$ and non-head parameters $\mathbf{w}_{\backslash h_k}$. We rewrite Eq. \ref{eq:forward} as:
 \begin{equation}
    \hat{y}_k = h_k (f_{\backslash h_k}(x_k)),
\end{equation}
where the $h_k$ and $f_{\backslash h_k}$ denote the head and non-head module, respectively. 
 In the classification task, the head module can be a fully-connection layer with parameter matrix $\mathbf{w}_h \in \mathbb{R}^{E\times C}$.   

 Parameter decoupling allows FL not to care about the content before the head module $h$, thus realizing the heterogeneity of input data and majority of models. Although it's not ``completely'' heterogeneous, we claim that such a compromise is acceptable. Firstly, the aligned parameters only account for a very small part of the model, which hardly prevents the participants from designing and keeping their own model structures private. Let's take the well-known ResNet \cite{resnet} as an example. For the smallest ResNet-18 model, only around 4.38\% (0.51M of 11.69M) parameters' size should be fixed. This percentage can be less than 1\% for larger models like ResNet-152. Secondly, FL methods that communicate at the embedding level is also considered to be private enough \cite{heGroupKnowledgeTransfer2020}. Our method does not even need to transmit embeddings but the embedding-related parameters, which means less risk of information leakage and more communication-efficiency.
 
\subsection{``Data-Free'' knowledge distillation}
FL methods based on knowledge distillation can improve the performance in non-\textit{i.i.d.} tasks, but they often require some shared data samples between clients, which is almost impossible in the data heterogeneity case. We propose ``data-free'' KD on FL to solve the problem. Specifically, we keep two head modules in client $k$, namely, global head $\overline{h}$ and local head $h_k$. At the end of every local training epoch, we aggregate all local heads' parameters $\mathbf{w}_{h_k}$ to the server and update the global head's parameter $\overline{\mathbf{w}}_h$:
\begin{equation}
    \overline{\mathbf{w}}_h = \sum_{k=1}^{K} \mathbf{w}_{h_k}.
\end{equation}

Client $k$ predict the probability from the local $\hat{y}_k$ and global head $\overline{y}_k$ from the second training epoch simultaneously. Let $\overline{y}_k$ be the fixed teacher signal within a training epoch. Then the loss of each client in Eq. \ref{eq:opt} can be modified as
\begin{equation}
\begin{aligned}
 &\mathcal{L}_k = \frac{1}{N_k} \sum_{n=1}^{N_k} \{\mathcal{L}_{\text{CE}}(\hat{y}_{k}^{n}, y_{k}^{n}) + \mathcal{L}_{\text{DKD}}(\hat{y}_{k}^{n}, \overline{y}_{k}^{n}, \mathcal{T})\},\\ 
 \text{where \;}&\hat{y}_k = h_k (f_{\backslash h_k}(x_k)), \text{\;}\overline{y}_k = \overline{h}_k (f_{\backslash h_k}(x_k)).
\end{aligned}
\end{equation}

In this way, we achieve KD without shared data across clients. We also introduce the improved Decoupled Knowledge Distillation (DKD) loss \cite{DKD} to mitigate the negative effects due to complete heterogeneity and non-\textit{i.i.d}.  
\begin{equation}
    \begin{aligned}
\mathrm{DKD}= \alpha[ \mathrm{KL}( {\overline{\mathbf{b}}} \| \hat{\mathbf{b}}) + \mathrm{KL}(\overline{\mathbf{p}} ;\mathcal{T}\| \hat{\mathbf{p}})],
\end{aligned}
\end{equation}
where $\alpha$ is hyper-parameters, KL is the Kullback-Leibler Divergence, $\mathbf{b}$ is the probability corresponding to the true label $y$, $\mathbf{p}$ is the re-normalized relative probability of the remaining classes, and $\mathcal{T}$ denotes temperature. Compared with vanilla DKD, we reduce the hyper-parameters to control the absolute strength of KD Loss. In addition, we introduce a temperature $\mathcal{T}$ related to current epoch $T$ to avoid global head $h$ producing overconfident signals on unreal labels in early stages of training. Let the maximum epoch be $T_{\max}$, we have
\begin{equation}
\begin{aligned}
    &\overline{\mathbf{p}} = \frac{\exp \left(\overline{z}_i / \mathcal{T}\right)}{\sum_j \exp \left(\overline{z}_j / \mathcal{T}\right)}, \\
    &\mathcal{T} = \beta\left(1 + \cos(\frac{T}{T_{\max} }\pi)\right)+1,
\end{aligned}
\end{equation}
where $z_i$ is the $i$-th logit (in $\overline{y}$) of each category output except the ground truth.
The pseudo-code of our method is summarized as Algorithm \ref{algorithm}.

\begin{algorithm}[]
  \SetAlgoLined\DontPrintSemicolon
  \SetKwFunction{algo}{a}\SetKwFunction{proc}{ClientUpdate}
  \SetKwProg{myalg}{Server executes:}{}{}
  \KwData{Client number $K$, datasets $\mathcal{D}_k$, private model $f_k$ with local heads $h_k$ (weight $\mathbf{w}_{h_k}$), global $\overline{h}$ (weight $\mathbf{w}_h$), the number data samples $N_k$, and total training epochs $T_{\max}$.}
  \myalg{}{
  initialize $\mathbf{w}_h$\;
    \For{$T=1,2,\cdots,T_{max}$}
    {
    \For{$k=1,2,\cdots,K$ \textbf{in parallel}}{
     \nl $\mathbf{w}_{h_k}$ =\proc{$k, \overline{h}, T$}\;
	}
	$\overline{\mathbf{w}}_h = \sum_{k=1}^{K} \mathbf{w}_{h_k}$\;
	}
  \nl \KwRet\;}{}
  \setcounter{AlgoLine}{0}
  \SetKwProg{myproc}{ClientUpdate ($k, \overline{h}, T$):}{}{}
  \myproc{{}}{
  $\mathcal{T} = \beta\left(1 + \cos(\frac{T}{T_{\max} }\pi)\right)+1$\;
 \For{$n=1,2,\cdots,N_k$ }{
    $\hat{y}_k^n = h_k (f_{\backslash h_k}(x_k^n))$\;
    $\overline{y}_k^n = \overline{h} (f_{\backslash h_k}(x_k^n))$\;}
    \eIf{$T \neq 1$}{$\mathcal{L}_k = \frac{1}{N_k} \sum_{n=1}^{N_k} \mathcal{L}_{\text{CE}}(\hat{y}_{k}^{n}, y_{k}^{n}) + \alpha \mathcal{L}_{\text{DKD}}(\hat{y}_{k}^{n}, \overline{y}_{k}^{n}, \mathcal{T})$ }
    {$\mathcal{L}_k = \frac{1}{N_k} \sum_{n=1}^{N_k} \mathcal{L}_{\text{CE}}(\hat{y}_{k}^{n}, y_{k}^{n}) $\;}
  \nl \KwRet $\mathbf{w}_{h_k}$\;}
  \caption{Completely Heterogeneous Federated Learning}
  \label{algorithm}
\end{algorithm} 

\begin{table}[]
\centering
\resizebox{1\linewidth}{!}{
\begin{tabular}{cccccc}
\toprule
Dataset                & \textit{I.I.D.} & SOLO            & AVG             & AVGKD           & Ours                             \\ \midrule
\multirow{2}{*}{DIGIT} & \CheckmarkBold & 85.77$\pm$0.76  & 78.16$\pm$3.63  & 75.45$\pm$5.27  & \textbf{86.46$\pm$0.89}   \\
                       & \XSolidBrush   & 79.04$\pm$3.06  & 64.48$\pm$9.64  & 63.55$\pm$9.16  & \textbf{82.87$\pm$1.65}   \\ \midrule
\multirow{2}{*}{ADULT} & \CheckmarkBold & 72.83$\pm$4.92  & 68.68$\pm$6.03  & 64.55$\pm$11.00 & \textbf{73.69$\pm$4.74}   \\
                       & \XSolidBrush   & 65.05$\pm$8.54  & 61.54$\pm$14.44 & 56.96$\pm$11.44 & \textbf{65.65$\pm$8.17}   \\ \midrule
\multirow{2}{*}{HCC}   & \CheckmarkBold & 66.85$\pm$10.19 & 66.82$\pm$6.29 & 68.79$\pm$4.97  & \textbf{73.83$\pm$8.11}   \\
                       & \XSolidBrush   & 59.23$\pm$19.28 & 59.56$\pm$18.68 & 58.30$\pm$16.65 & \textbf{61.39$\pm$17.20}            \\ \midrule
\multirow{2}{*}{BCD}   & \CheckmarkBold & 81.85$\pm$11.89 & 82.65$\pm$11.91 & 87.67$\pm$12.02  & \textbf{89.88$\pm$10.08}    \\
                       & \XSolidBrush   & 71.40$\pm$19.20 & 78.56$\pm$10.73 & 83.31$\pm$14.38 & \textbf{90.66$\pm$7.07}   \\ \midrule
\multirow{2}{*}{ILPD}  & \CheckmarkBold & 61.87$\pm$13.64 & 59.07$\pm$11.95 & 62.99$\pm$9.92  & \textbf{63.79$\pm$10.66}    \\
                       & \XSolidBrush   & 53.32$\pm$15.49 & 54.42$\pm$20.08 & \textbf{58.02$\pm$19.05} & 55.18$\pm$16.70    \\ \bottomrule
\end{tabular}
}
\caption{Top-1 accuracy comparison on different datasets. \textbf{Bold} denotes the highest accuracy.}
\label{table:class}
\end{table}
\section{Experiments}
  We compare different strategies across five image and tabular datasets with heterogeneous distribution, feature domain, and model architecture. Heterogeneous distribution refers to the non-\textit{i.i.d.} labels, specifically the Dirichlet distribution (alpha is 0.5). The experimental results show our method's generalizability under completely heterogeneous scenarios.

\subsection{Image datasets and models}
\begin{figure}
   \centering 
    \subfigure[MNIST (Gray-scale)]{
        \label{Fig.digit.1}
        \includegraphics[width=0.305\linewidth]{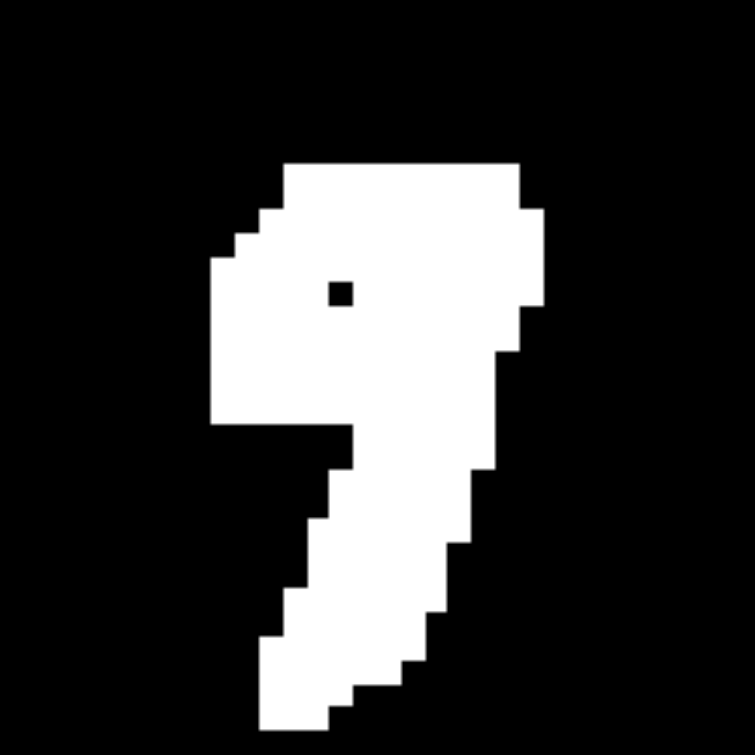}
        }
    \subfigure[MNIST-M (RGB)]{
        \label{Fig.digit.2}
        \includegraphics[width=0.305\linewidth]{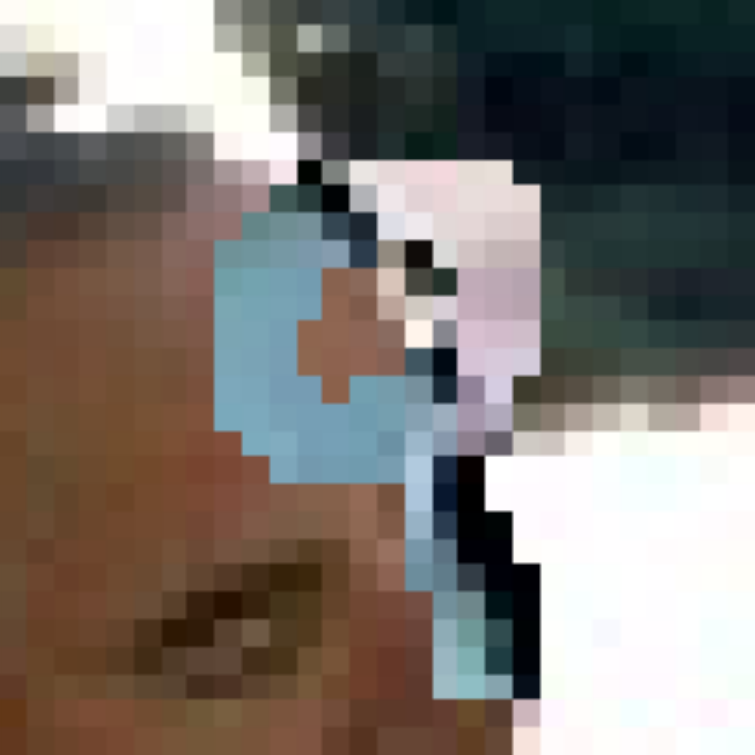}
        }
    \subfigure[SYN (RGB)]{
        \label{Fig.digit.3}
        \includegraphics[width=0.305\linewidth]{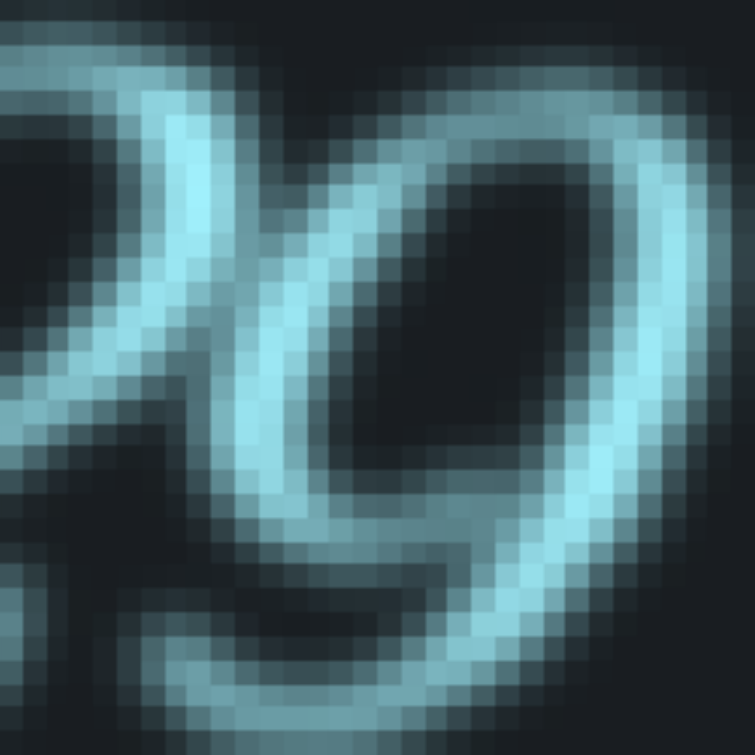}
        }

    \caption{Number 9 in heterogeneous datasets.}
    \label{Fig.digit}
    \centering 
\end{figure}

To simulate the completely heterogeneous scenarios, we introduce \textbf{MNIST} \cite{MNIST},  \textbf{MNIST-M} ($1\times28\times28$ gray-scale images) \cite{MNIST-M} ($3\times32\times32$ RGB images), and \textbf{Synthetic Digits} ($3\times32\times32$ RGB images) \cite{MNIST-M} as our test datasets. Fig. \ref{Fig.digit} illustrates the number 9 in different datasets. We choose the heterogeneous MobileNet-V3 \cite{mobilenet}, ResNet-18 \cite{resnet}, and VGG \cite{vgg} for each dataset respectively.

\subsection{Tabular datasets and models}
\begin{figure}[]
  \centering
  \includegraphics[width=0.9\linewidth]{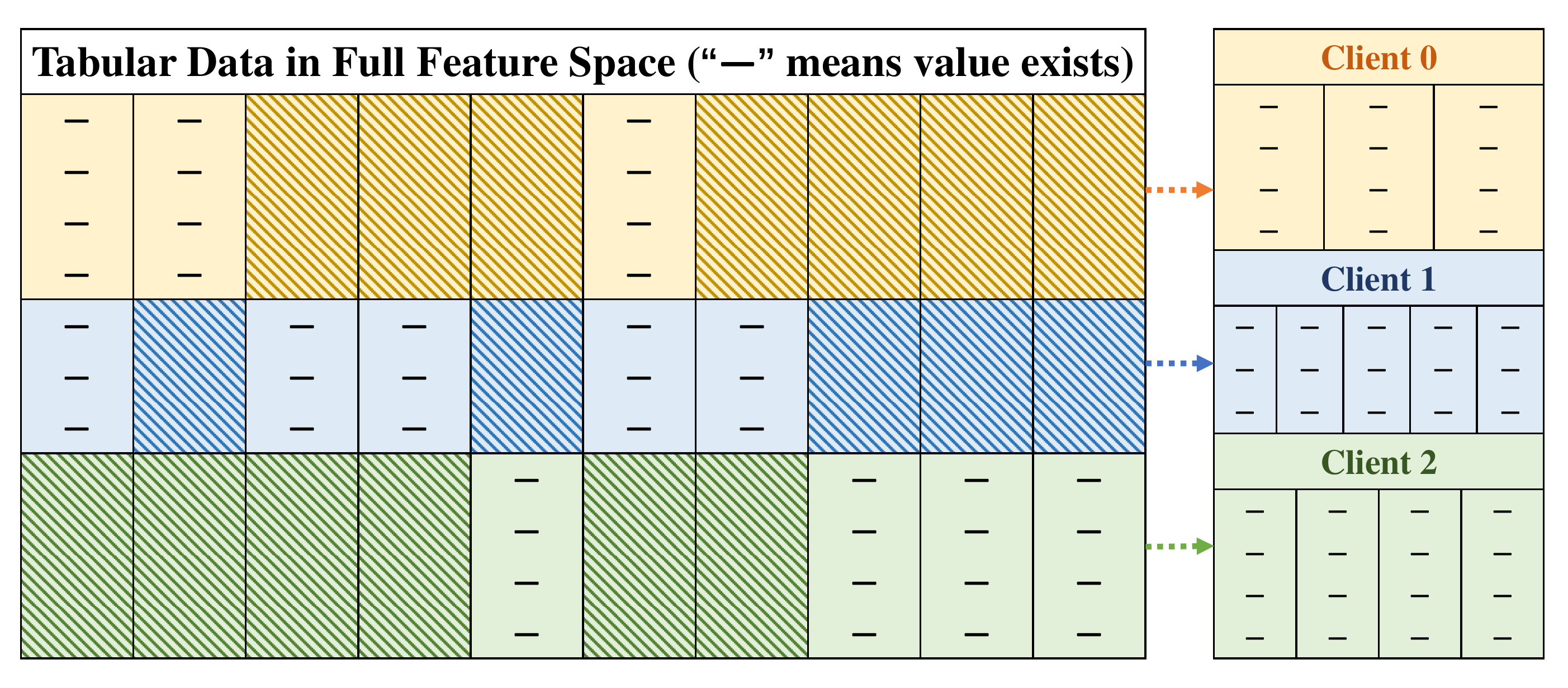}
\caption{Splitting heterogeneous data on tabular datasets.}
\label{fig:tabdata}
\end{figure}

We choose the \textbf{ADULT} \cite{UCI}, \textbf{HCC} \cite{HCC}, \textbf{BCD} \cite{BCD} and \textbf{ILPD} \cite{UCI} datasets to verify the effect in the tabular scenario. We first slice each dataset into sub-datasets with different features \cite{SubTab} and sample IDs \cite{FedLab} and submit them to each client, as shown in Fig. \ref{fig:tabdata}, and then divide the training and testing sets within the client. Then, we generate MLP models randomly with different layers and hidden unit numbers on each client.  

\subsection{Fair comparison and result analysis}
We compare our method (\textbf{Ours}) with three training strategies, \textit{i.e.,} \textbf{SOLO}: each client trains locally, \textbf{AVG}: server aggregates average weight for model head after each training epoch, which is the same as existing parameter decoupling methods \cite{arivazhaganFederatedLearningPersonalization2019,collinsExploitingSharedRepresentations2021, FedPS}, and \textbf{AVGKD}: training with our method and AVG at the same time. 
We record the best average validation accuracy of all clients each time. We repeat each experiment on five random seeds and compute the mean and standard deviation of accuracy without cherry-picking. 
To further ensure a fair comparison, we use the same number of training epochs, and adjust the hyper-parameters (such as learning rate and batch size) to make each client fully converge. All clients' training accuracy reaches 100\%. The hyper-parameters of the four training strategies are consistent during training on the same dataset. The hyper-parameters of our proposed method (\textit{i.e.,} $\alpha,\beta$ in Algorithm \ref{algorithm}) are constant at 0.5 and 5.

As shown in Table. \ref{table:class}, our method obtains the best accuracy in most cases. We also listed the effect of the method when the data is \textit{i.i.d.} distributed, where \textit{I.I.D.} column shows \CheckmarkBold in table.
In most cases, the simultaneous heterogeneity of data features and models harms federated learning performance greatly. The existing decoupling method (AVG) is difficult to handle the completely heterogeneous scenarios. When the data categories distribution is also heterogeneous (non-\textit{i.i.d.}), federated learning will be further adversely affected. \textbf{Our method is the only one that can guarantee positive effects in all cases}, since all of the other strategies may be even lower than just local training.

\section{Conclusion}
This paper models a new ``completely heterogeneous'' scenario for FL, where the feature, data distribution, and model are all heterogeneous. 
We tackle this challenge by proposing parameter decoupling and data-free knowledge distillation. Parameter decoupling enables clients to keep most of their private model secrete and make FL possible across different feature spaces. Data-free knowledge distillation means clients don't need to share any data samples with other devices and worry about how to align heterogeneous feature spaces. Moreover, we modify the decoupled knowledge distillation loss to improve the performance when the data is non-\textit{i.i.d.} Experiments illustrate the superior performance of the proposed training methodology. 
\vfill\pagebreak

\clearpage


\bibliographystyle{IEEEbib}
\small
\bibliography{refs}

\end{document}